%% file: main.tex
\documentclass[conference]{IEEEtran}
\IEEEoverridecommandlockouts
\usepackage{cite}
\usepackage{amsmath,amssymb,amsfonts}
\usepackage{algorithmic}
\usepackage{graphicx}
\usepackage{textcomp}
\usepackage{xcolor}
\usepackage{hyperref}
\usepackage{booktabs}
\usepackage{adjustbox}
\def\BibTeX{{\rm B\kern-.05em{\sc i\kern-.025em b}\kern-.08em
    T\kern-.1667em\lower.7ex\hbox{E}\kern-.125emX}}
\begin{document}

\title{Multi-Focus Temporal Shifting for Precise Event Spotting in Sports Videos}

\author{
    \IEEEauthorblockN{Hao Xu\IEEEauthorrefmark{1}, Xinyu Wei\IEEEauthorrefmark{2}, Sam Wells\IEEEauthorrefmark{3}, and Sunil Aryal\IEEEauthorrefmark{1}}
    \IEEEauthorblockA{\IEEEauthorrefmark{1}Deakin University, Melbourne, Australia}
    \IEEEauthorblockA{\IEEEauthorrefmark{2}Champion Data, Melbourne, Australia}
    \IEEEauthorblockA{\IEEEauthorrefmark{3}Paralympics Australia, Melbourne, Australia}
    \IEEEauthorblockA{\texttt{\small august.xu@research.deakin.edu.au, sunil.aryal@deakin.edu.au}}
    \IEEEauthorblockA{\texttt{\small felix.wei@championdata.com.au, sam.wells@paralympic.org.au}}
}
\maketitle
\input{sec/0_abstract}
\input{sec/1_intro}

\input{sec/2_related_work}
\input{sec/3_methods}
\input{sec/4_emprical_study}
\input{sec/6_conclusion}

{
    \small
    \bibliographystyle{IEEEbib}
    \bibliography{main}
}

\end{document}

%% file: sec/0_abstract.tex
\begin{abstract}
Precise Event Spotting (PES) in sports videos requires frame-level recognition of fine-grained actions from single-camera footage. Existing PES models typically incorporate lightweight temporal modules such as the Gate Shift Module (GSM) or the Gate Shift Fuse to enrich 2D CNN feature extractors with temporal context. However, these modules are limited in both temporal receptive field and spatial adaptability. 
We propose Multi-Focus Temporal Shifting Module (MFS) that enhances GSM with multi-scale temporal shifts and Group Focus Module, enabling efficient modeling of both short and long-term dependencies while focusing on salient regions. MFS is a lightweight, plug-and-play module that integrates seamlessly with diverse 2D backbones.
To further advance the field, we introduce the Table Tennis Australia dataset, the first PES benchmark for table tennis containing over 4,800 precisely annotated events. 
Extensive experiments across five PES benchmarks demonstrate that MFS consistently improves performance with minimal overhead, achieving leading results among lightweight methods (+4.09 mAP, 45 GFLOPs).
All code and datasets are publicly available at \href{https://github.com/AugustRushG/MFS}{https://github.com/AugustRushG/MFS}.
\end{abstract}

%% file: sec/1_intro.tex
\section{Introduction}
\label{sec:intro}
Detecting the precise moment when events occur in a video is a core challenge in video understanding, particularly in the single-camera setting commonly used in most sports recordings. Precise Event Spotting (PES) aims to localize and classify events at frame-level accuracy, typically within a 1–2 frame tolerance \cite{hong2022spotting}. This fine-grained precision is critical in sports scenarios, where even a few frames of deviation can correspond to a different event entirely. Accurate PES provides a strong foundation for downstream tasks such as ball tracking, player positioning, and automatic highlight generation.
Moreover, precise spotting plays a vital role in post-game analysis by highlighting key moments and reducing the need to manually browse full-length videos. For instance, according to Table Tennis Australia’s sports analytics team, annotating events in a single match video can take 4–5 hours. Automating this process not only streamlines analysis but also enables timely feedback for athletes and coaches.

\begin{figure}
    \centering
    \includegraphics[width=1\linewidth]{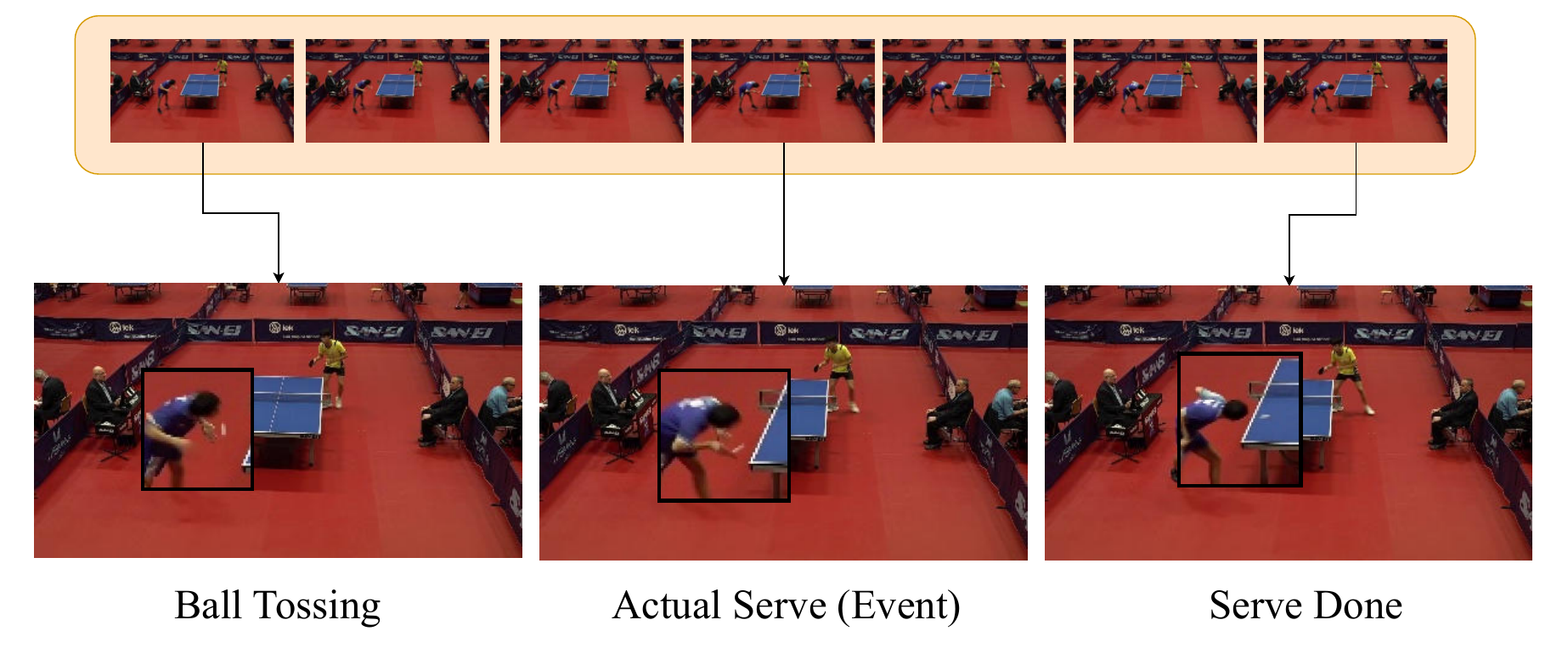}
    \caption{Illustration of a table-tennis serve sequence. Identifying the actual serve event requires information from earlier frames showing the ball toss and preparation, as well as later frames showing the completion phase. This demonstrates that temporal cues beyond t±1 are necessary for accurate event spotting.}
    \label{fig:teaser}
\end{figure}

\begin{figure}
    \centering
    \includegraphics[width=1\linewidth]{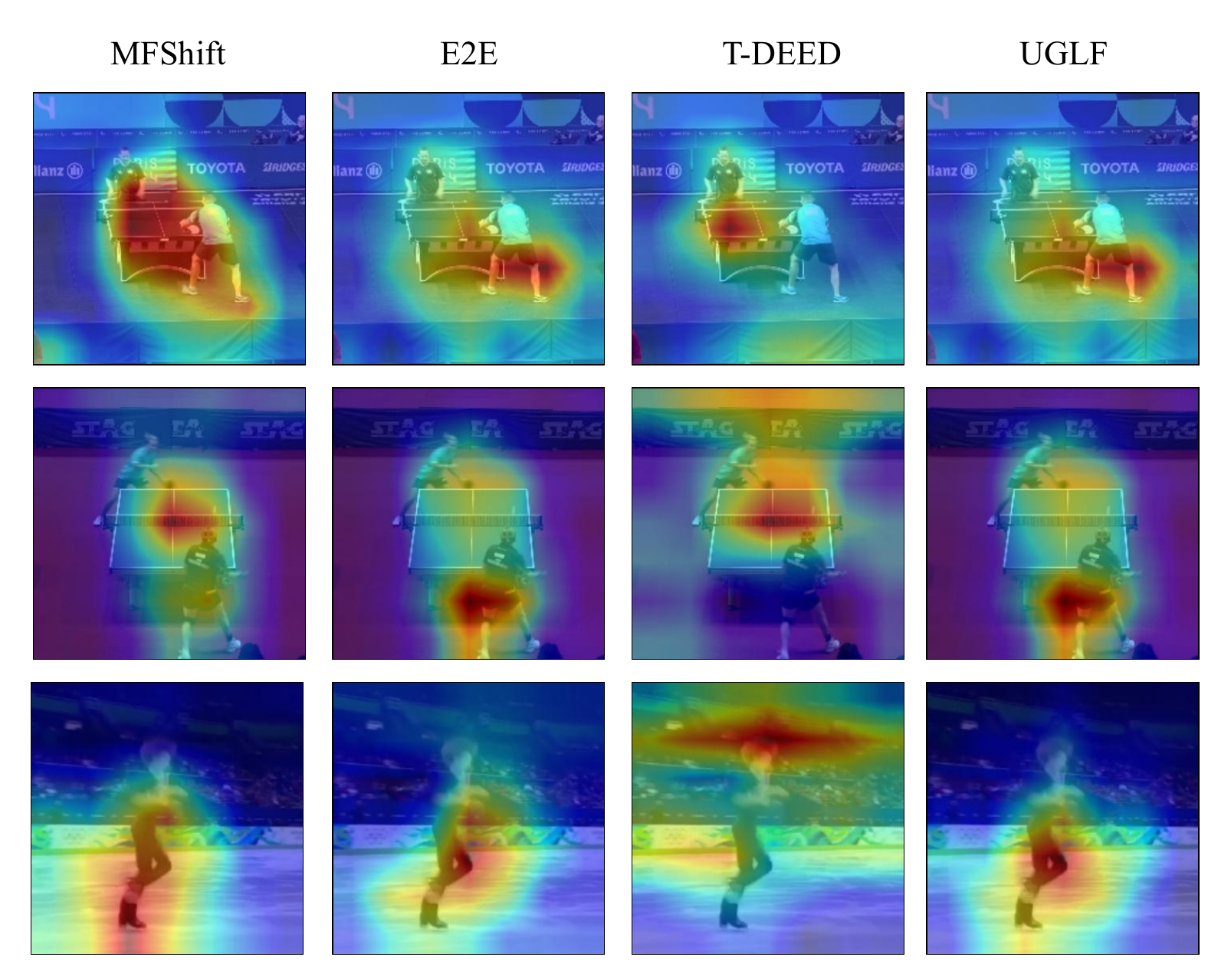}
    \caption{Heatmap visualizations across three event types:
Row 1 — occluded ball-bounce; Row 2 — normal ball-bounce; Row 3 — jump-landing.
MFS yields sharper, more meaningful focus regions, attending to players/table when the ball is occluded, to ball–racket interactions during normal bounce, and to the skater’s boot at landing.}
    \label{fig:attention_difference}
\end{figure}

Recent progress in video understanding has introduced transformer-based models \cite{cao2022spotformer, zhu2022transformer}, multimodal architectures \cite{vanderplaetse2020improved, xarles2023astra}, and self-supervised frameworks \cite{denize2024comedian}.
E2E-Spot \cite{hong2022spotting} remains a widely adopted baseline for PES and has inspired several state-of-the-art (SOTA) extensions that incorporate increasingly heavy temporal modules, including T-DEED \cite{xarles2024t}, UGLF \cite{tran2024unifying}, and ASTRM \cite{santra2025precise}.
At the core of these lightweight models lies the Gate Shift Module (GSM) \cite{sudhakaran2020gate}, which enriches 2D CNNs with temporal cues through adjacent-frame shifting. 
However, GSM has two key limitations: a restricted temporal receptive field (t±1), which works only for near-instantaneous events and fails when longer temporal context is required (Figure~\ref{fig:teaser}), and a lack of spatial selectivity (Figure~\ref{fig:attention_difference}), which limits meaningful temporal exchange when salient regions occupy only a small portion of the frame.
While recent works like ASTRM \cite{santra2025precise} address these issues using heavy temporal modules, they incur high computational costs. We argue that complex architectures are not required. By applying principled multi-scale shifts and grouped focusing, we achieve competitive performance while maintaining 
practical efficiency for deployment.

PES remains challenging because events often differ from their surrounding frames only subtly—sometimes requiring several frames of context for human annotators to disambiguate \cite{xarles2024t}. Different event types also rely on varying temporal ranges, making fixed-range temporal modeling insufficient. Furthermore, the dominance of background pixels in sports videos requires models to attend to small but salient regions (e.g., ball contact points), highlighting the need for efficient spatial focusing in addition to temporal reasoning.
Another practical challenge is the lack of suitable datasets. Most public sports video datasets are taken from broadcast footage \cite{hong2022spotting, hong2021video, giancola2018soccernet, xu2022finediving}, recorded from stable, high-quality camera views. In contrast, real-world settings—such as local or developmental leagues—often rely on handheld recordings with varied viewpoints, inconsistent quality, and frequent occlusions. Existing datasets also tend to have sparse event distributions, making them less representative of sports with rapid, densely packed actions.


To address the identified limitations in temporal modeling, we propose Multi-Focus Temporal Shifting (MFS), a lightweight extension of GSM that addresses both temporal range and spatial selectivity through principled design choices. 
MFS captures long-range temporal dependencies through multi-scale feature shifts rather than fixed adjacent-frame operations, providing direct access to broader temporal windows (e.g., t±1, t±3, t±6) when needed. To manage the increased noise from larger temporal windows, we introduce a Grouped Focus Module (GFM) that divides feature channels into groups, each learning independent spatial attention masks to focus on distinct salient regions (e.g., ball position, player movements). 
Unlike transformer-based multi-head attention which requires expensive query–key–value computations, GFM provides efficient spatial selectivity through simple grouped convolutions.

Beyond our methodological contributions, we also fill a critical gap in PES benchmarks by introducing the Table Tennis Australia (TTA) dataset. Unlike existing datasets captured from broadcast-quality recordings of elite competitions, TTA contains realistic footage collected using handheld cameras with varied viewpoints, motion blur, and frequent occlusions—closely mirroring real-world sports analytics scenarios under limited-budget conditions. With 4,878 annotated events across 39 matches, TTA exhibits approximately \textbf{2$\times$} higher event density than prior benchmarks, presenting unique challenges for fine-grained temporal modeling. The dataset will be made publicly available for research use.

\noindent\textbf{Our main contributions are as follows:}
(1) We propose \textbf{MFS}, a plug-and-play module that integrates multi-scale temporal shifting with grouped spatial focusing, achieving superior performance with minimal computational overhead (1.3\% additional parameters);
(2) We introduce the \textbf{TTA} dataset—the first PES benchmark for table tennis—featuring realistic recording conditions and 2× higher event density than existing benchmarks, addressing critical gaps in available resources; and
(3) Through extensive experiments across five benchmarks and three architectural frameworks, we demonstrate that MFS consistently advances SOTA methods by up to 4.09\% while maintaining efficiency, achieving leading performance among lightweight approaches with only 45 GFLOPs compared to 67.24 GFLOPs for recent complex architectures.

%% file: sec/2_related_work.tex
\section{Related Work}
Video understanding tasks such as classification and action recognition typically rely on \textit{sparse} frame sampling \cite{feichtenhofer2019slowfast, wang2016temporal, kondratyuk2021movinets}. In contrast, Precise Event Spotting (PES) requires \textit{dense} sampling to localize fine-grained events where visual cues are subtle and temporally narrow \cite{xu2025action}. Unlike Temporal Action Detection \cite{lin2018bsn, su2021bsn++}, which predicts action intervals, PES aims to pinpoint the exact event frame.

Two main paradigms exist for temporal modeling in PES: transformer-based approaches and lightweight shift-based modules. Transformers \cite{cao2022spotformer, zhu2022transformer, xarles2023astra} or complicated temporal modules \cite{santra2025precise} capture long-range context but incur high computational cost (600+ GFLOPs), whereas lightweight methods such as TSM \cite{lin2019tsm}, GSM \cite{sudhakaran2020gate}, and GSF \cite{sudhakaran2023gate} deliver competitive accuracy with far lower overhead (40–45 GFLOPs). These shift-based modules form the foundation of several SOTA PES models \cite{hong2022spotting, xarles2024t, tran2024unifying}.

The PES task emerged from SoccerNet \cite{giancola2018soccernet}, which evolved from relaxed ($\pm$50 f) to high-precision tolerances. E2E-Spot \cite{hong2022spotting} established an efficient baseline using GSM with RegNetY. Subsequent works improved temporal discrimination via finer-grained perception (T-DEED \cite{xarles2024t}), vision–language fusion (UGLF \cite{tran2024unifying}), or spatial–temporal feature injection and contrastive learning (ASTRM \cite{santra2025precise}). Multimodal approaches such as ASTRA \cite{xarles2023astra} further incorporate audio cues.

Despite the strong performance of GSM-based methods, GSM suffers from two core limitations: (1) a fixed adjacent-frame shift (t±1), restricting the temporal receptive field, and (2) a lack of spatial selectivity, even though sports videos contain large background regions. Rather than introducing a new architecture, we extend the efficiency of shift-based modeling. Our proposed \textbf{MFS} addresses these limitations using multi-scale temporal shifts and grouped spatial focus, achieving leading accuracy among lightweight methods (45 GFLOPs).

%% file: sec/3_methods.tex
\section{Method}
\subsection{Overview and Design Rationale}
We introduce two targeted extensions: (1) \textbf{multi-scale temporal shifts} to access longer temporal ranges, and (2) \textbf{grouped spatial focusing} for efficient region-aware modulation.  
MFS integrates into standard 2D CNN backbones as a drop-in replacement for GSM. These additions introduce only 0.06M parameters (1.3\% overhead) while providing broader temporal receptive fields and spatial selectivity.

\subsection{Grouped Focus Module (GFM)}
In sports videos, event-critical information is localized to small regions—ball position, player hand movements, or racket contact—while 60-80\% of pixels represent background. A single global attention map forces all feature channels to compete for the same spatial focus, limiting the ability to simultaneously track multiple relevant entities (e.g., ball \textit{and} player position).
To enable efficient multi-region attention, GFM divides feature channels into $G$ groups (default $G\!=\!2$), each learning an independent spatial focus map. This allows different groups to specialize on distinct spatial regions without competition. For each group $F_g \in \mathbb{R}^{C/G\times H \times W}$ (where $C$, $H$ and $W$ denote channels, height and width, respectively), a lightweight 3 $\times$ 3 convolution followed by sigmoid activation generates a focus map:

\begin{equation}
\hat{F}g = F_g \odot \sigma\left(\mathrm{Conv}{3\times3}(F_g)\right)
\end{equation}

\begin{figure}
    \centering
    \includegraphics[width=0.9\linewidth]{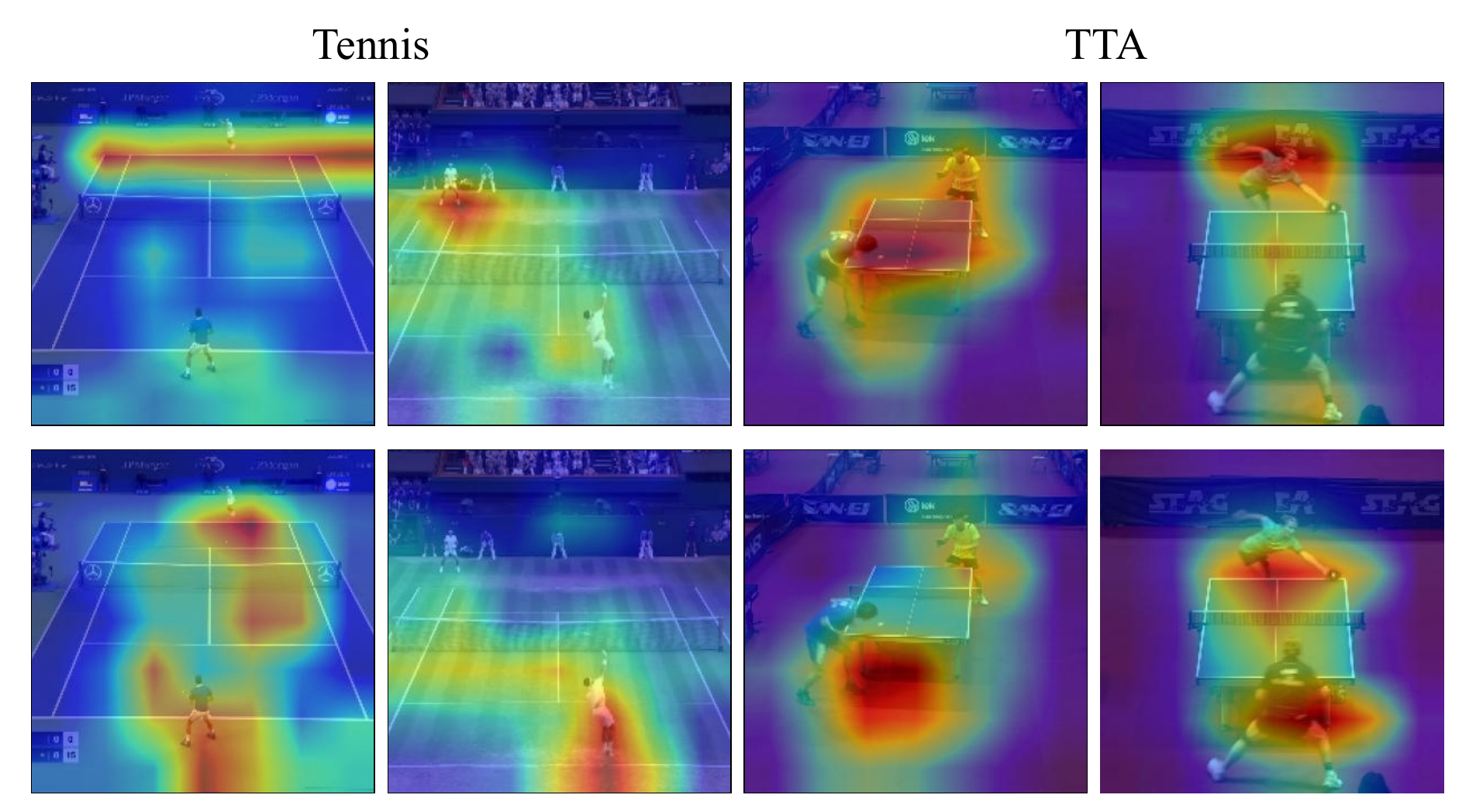}
    \caption{Grad-CAM heatmaps on Tennis and TTA. Row 1 (Group 1) and Row 2 (Group 2) attend to different event-critical regions—e.g., the two players in Tennis, and the table versus the ball in TTA—showing that grouped focusing produces complementary spatial attention.}
    \label{fig:group_focus}
\end{figure}

Where $\odot$ denotes element-wise multiplication. All focused groups are concatenated along the channel dimension to form the refined feature representation $F' = [\hat{F}_1, \hat{F}_2, \ldots, \hat{F}_G]$. 
By processing groups independently, each can specialize—one might attend to the ball, another to player positions. This multi-focus design improves robustness: if one region is occluded or blurred, others remain informative. Compared to transformer multi-head attention which requires expensive query-key-value computations, GFM achieves spatial selectivity through simple grouped convolutions, adding minimal overhead while enabling simultaneous attention to multiple event-critical regions.
To further demonstrate the spatial selectivity learned by GFM, we provide Grad-CAM–based heatmap visualizations in Figure \ref{fig:group_focus}. These examples show how MFS attends to semantically meaningful regions across diverse event types.

\subsection{Multi-Scale Shift Module}
\begin{figure}[t]
    \centering
    \includegraphics[width=0.9\linewidth]{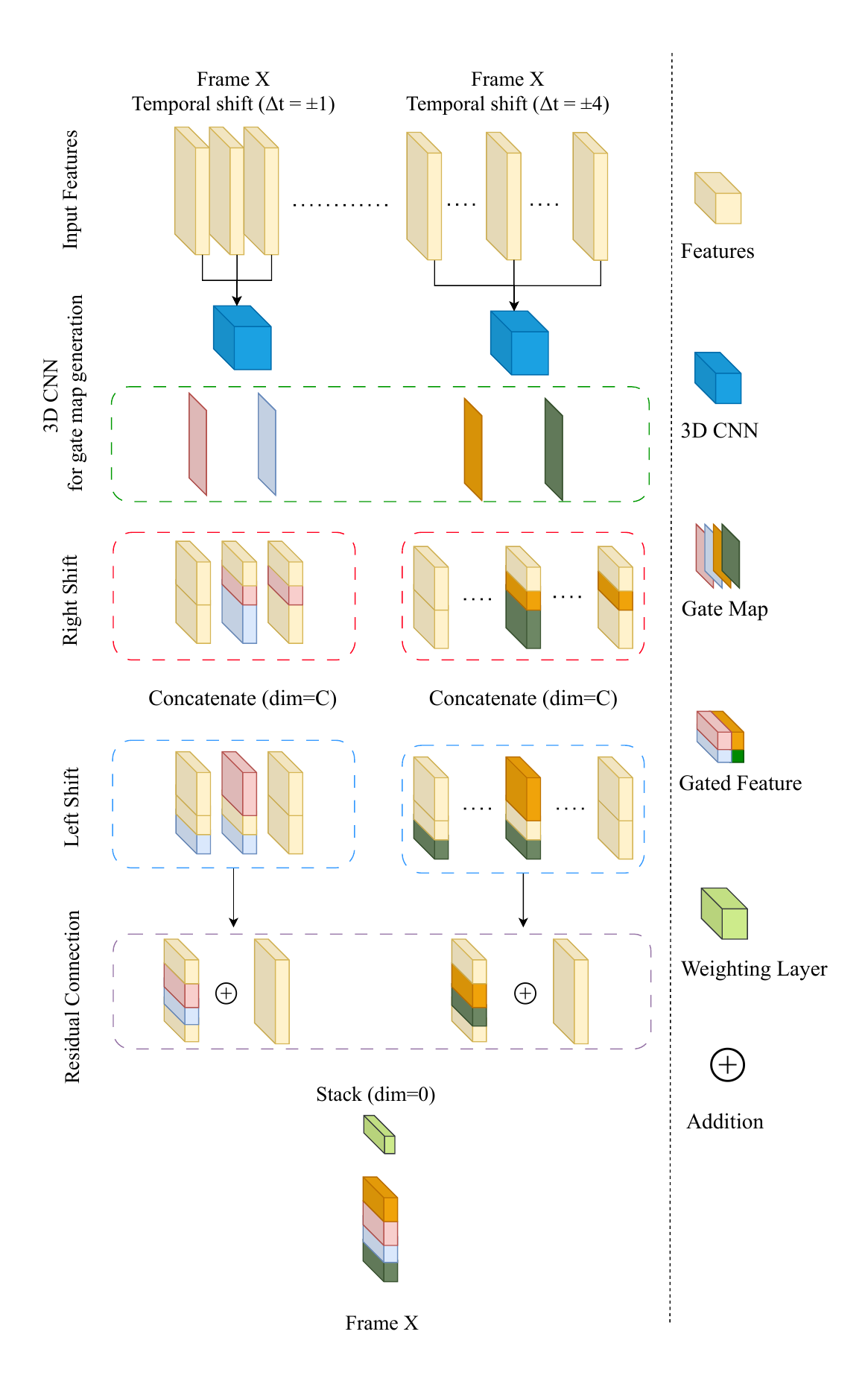}
    \caption{Multiscale Shift overview. A 3D CNN generates gate maps for different temporal shift ranges (e.g., $\Delta t = \pm1, \pm3$). Features are shifted bidirectionally across multiple scales, gated, and fused through learnable weights, enabling adaptive capture of both short and long-term temporal dependencies.}
    \label{fig:msgsm}
\end{figure}

GSM has proven effective in PES tasks \cite{hong2022spotting, xarles2024t, tran2024unifying}, performing gating-based temporal shifts inspired by TSM \cite{lin2019tsm} and GST \cite{luo2019grouped}. However, its fixed adjacent-frame operation ($t\pm1$) limits access to broader temporal context. Analysis of event durations in the Tennis dataset reveals that 73\% of events span 3--8 frames, beyond GSM's receptive field. Events such as serves or shot preparation require this extended context for accurate recognition, motivating the need for multi-scale temporal access.
MFS extends GSM by introducing learnable multi-scale temporal shifts, providing direct access to frames at varying distances (e.g., $\delta \in \{1, 2, 3\}$). Figure~\ref{fig:msgsm} illustrates the architecture.
Given feature maps $X \in \mathbb{R}^{C \times T \times H \times W}$ (where $T$ denotes the temporal length), MFS applies gated temporal shifts at multiple distances. For each shift distance $\delta$, a 3D convolutional gating layer processes a three-frame temporal window to generate left and right gates:
\begin{equation}
(G_{\text{left}}, G_{\text{right}}) = \tanh(\text{Conv3D}(X))
\end{equation}
The feature map is split into two halves, each modulated by its gate and shifted temporally:
\begin{align}
\hat{X}_{\text{left}} &= \text{Roll}(X_{\text{left}} \odot G_{\text{left}}, -\delta) \\
\hat{X}_{\text{right}} &= \text{Roll}(X_{\text{right}} \odot G_{\text{right}}, +\delta)
\end{align}
where $\odot$ denotes element-wise multiplication and $\delta$ is the temporal shift distance in frames. Zero padding maintains sequence length at boundaries. Shifted features are combined with a residual connection:
\begin{equation}
\hat{X} = [\hat{X}_{\text{left}}, \hat{X}_{\text{right}}] + X
\end{equation}
Outputs from different shift distances $\{\hat{X}^{(i)}\}_{i=1}^{N}$ are combined through learnable weights, all of which are initialized to 1:
\begin{equation}
\hat{X}_{\text{final}} = \sum_{i=1}^{N} w_i \cdot \hat{X}^{(i)}, \quad \text{where } \sum_{i=1}^{N} w_i = 1,\; w_i \geq 0
\end{equation}
where $\{w_i\}$ are normalized via softmax. This allows the model to adaptively weight temporal scales—emphasizing nearby frames for rapid events (e.g., ball bounces at $\delta=1$) or distant frames for extended events (e.g., serves at $\delta=3$).

Multi-scale shifts expand temporal context with minimal complexity, while learnable weights let the model adaptively select the most relevant scales for each event. Combined with GFM’s spatial focus, MFS attends to the right regions at the right times. Despite adding only 2.7 GFLOPs (6\% overhead), it substantially enlarges the temporal receptive field compared to GSM.

%% file: sec/4_emprical_study.tex


\section{Empirical Study}
\subsection{Datasets}
We evaluate MFS on five fine-grained sports video datasets with frame-level annotations: 
\textbf{Tennis} \cite{zhang2021vid2player} contains 3,345 clips from 28 matches (25--30 FPS) with 33,791 frame-accurate annotations across six event classes.
\textbf{Figure Skating (FS)} \cite{hong2021video} includes 11 broadcast videos (25 FPS) covering 371 short programs from international competitions (2010--2019), with 3,674 annotated events across four classes. Following prior work \cite{hong2021video, xarles2024t, tran2024unifying}, we use the competition (\textbf{FS-Comp}) and performance (\textbf{FS-Perf}) splits.
\textbf{FineDiving} \cite{xu2022finediving} provides 3,000 clips with 7,010 events across four classes, converted to frame-level labels using the protocol of \cite{hong2022spotting}.
\textbf{Table Tennis Australia (TTA).} We introduce the first PES benchmark for table tennis, addressing critical gaps in existing datasets. TTA contains 39 full-game videos (30 FPS, 1280$\times$720 resolution) with 4,878 precisely annotated events across eight classes: serve, bounce, forehand, and backhand—each labeled separately for near and far table sides. The dataset is split into train/val/test sets with 3,759/598/521 events, ensuring diverse coverage across viewpoints and tournaments. Detailed annotation protocols are provided in \textbf{Supp.~§2}.
\begin{figure}[htbp]
\centering
\includegraphics[width=0.8\linewidth]{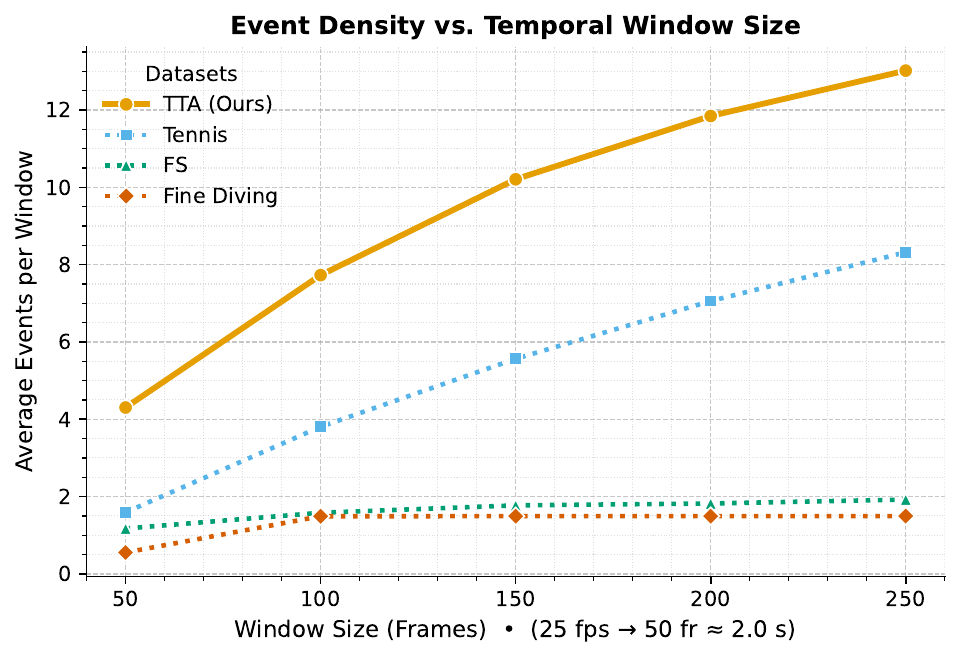}
\caption{Comparison of event density (average number of events) across varying temporal window sizes. TTA shows the highest event density across all ranges, reflecting its fine-grained and fast-paced nature.}
\label{fig:event-density}
\end{figure}

\begin{table*}[t]
\centering
\caption{Performance comparison (mAP \%) at different temporal tolerances ($\delta = 0, 1, 2$) across five fine-grained sports video datasets. The best result in each column is shown in \textbf{bold}, and our proposed method is \textit{italicized}. A $\star$ marks the best result within each model–backbone group. All models are evaluated with a fixed clip length of 100 frames.}
\label{tab:main-results}
\renewcommand{\arraystretch}{1.0}
\LARGE
\begin{adjustbox}{width=\textwidth}
\begin{tabular}{l c |ccc |ccc |ccc |ccc |ccc |c}
\toprule
\textbf{Model} & \textbf{Temp} &
\multicolumn{3}{c}{\textbf{TTA}} &
\multicolumn{3}{c}{\textbf{Tennis}} &
\multicolumn{3}{c}{\textbf{FS\_Comp}} &
\multicolumn{3}{c}{\textbf{FS\_Perf}} &
\multicolumn{3}{c}{\textbf{FineDiving}} &
\textbf{P (M) / GFLOPs}\\
& &
$\delta$=0 & $\delta$=1 & $\delta$=2 &
$\delta$=0 & $\delta$=1 & $\delta$=2 &
$\delta$=0 & $\delta$=1 & $\delta$=2 &
$\delta$=0 & $\delta$=1 & $\delta$=2 &
$\delta$=0 & $\delta$=1 & $\delta$=2 &
\\
\midrule
ASTRA\cite{xarles2023astra}& -  & 28.12& 54.16& 60.95& 49.98& 88.67& \textbf{97.49}& 35.87& 73.45& 84.71& 39.35& 78.10& 90.05& 27.28& 65.01& 82.74& 26.44 / 46.55\\
VIT \cite{dosovitskiy2020image} & -  & 14.17& 30.61& 43.38& 46.93& 83.71& 91.62& 33.59& 69.21& 83.73& 41.14& 77.47& 93.42& 27.70& 65.43& 83.30& 92.89 / 3522\\
ASTRM \cite{santra2025precise} & -  & 34.73 & 64.86 & 74.58 & 55.31 & 90.79 & 97.40 & 39.62 & 77.82 & 90.22 & 43.73 & 81.43 & 94.50 & \textbf{35.56} & 70.29 & 87.11 & 8.68 / 67.24 \\
\midrule
E2E-Spot\cite{hong2022spotting} & GSM     & 34.36& 66.42 & 76.90 & 51.42
& 90.03 & 96.48 & 36.44& 75.39 & 89.33 & 39.32& 81.27 & 94.14 & 27.31& 67.89 & 86.97 & 4.45 / 42.18\\
E2E-Spot & GSF     & 36.17& 63.70 & 72.19 & 49.68
& 89.01 & 96.55$^\star$ & 36.25& 75.49 & 89.32 & 40.07& 81.55 & 94.02 & 29.41& 67.57 & 86.22 & 4.45 / 42.20\\
\textit{E2E-Spot} & \textit{MFS} & \textit{\textbf{36.43}}$^\star$& \textbf{\textit{69.50}}$^\star$ & \textbf{\textit{78.12}}$^\star$ & \textit{\textbf{55.51}}$^\star$
& \textbf{\textit{91.34}}$^\star$ & \textit{96.51} & \textit{38.34}$^\star$ & \textit{77.72}$^\star$ & \textit{90.04}$^\star$ & \textit{46.54}$^\star$ & \textbf{\textit{84.12}}$^\star$ & \textbf{\textit{96.90}}$^\star$ & \textit{29.83}$^\star$ & \textit{\textbf{71.03}}$^\star$ & \textit{\textbf{87.26}}$^\star$ & \textit{4.51 / 45.06}\\
\midrule
T-DEED & GSM & 33.32$^\star$& 60.69 & 69.30 & 48.82
& 89.21 & 97.15$^\star$ & 38.61& 71.69 & 90.02 & 42.53& 79.62& 90.36& 26.86& 66.18 & 84.02 & 16.42 / 43.34\\
T-DEED\cite{xarles2024t} & GSF & 26.27& 54.04 & 64.51 & 50.08
& 89.58 & 97.06 & 38.83& 75.52 & \textbf{92.15}$^\star$ & 38.89& 79.82 & 94.79$^\star$ & 28.50& 67.04$^\star$ & 83.24 & 16.42 / 43.34 \\
\textit{T-DEED} & \textit{MFS} & \textit{28.06}& \textit{60.77}$^\star$& \textit{72.50}$^\star$& \textit{52.76}$^\star$
& \textit{90.68}$^\star$ & \textit{96.67} & \textit{39.06}$^\star$& \textit{75.55}$^\star$ & \textit{87.61} & \textit{44.66}$^\star$& \textit{82.41}$^\star$ & \textit{94.12} & \textit{26.88}$^\star$& \textit{66.70} & \textit{84.60}$^\star$ & \textit{16.48 / 46.20}\\
\midrule
UGLF\cite{tran2024unifying} & GSM & 34.36& 63.52 & 71.16 & 50.31
& 89.73 & 97.16$^\star$ & 36.99& 77.25 & 90.44 & 38.29& 79.24 & 93.08 & 27.11& 66.88 & 84.94 & 4.45 / 42.18\\
UGLF & GSF & 36.17 $^\star$& 63.70 & 72.19 & 50.94
& 89.39 & 97.06 & 34.84& 72.61 & 86.90 & 37.41& 82.06 & 94.38 & 29.81& 66.65 & 84.27 & 4.45 / 42.20\\
\textit{UGLF} & \textit{MFS} & \textit{33.93}& \textit{64.82}$^\star$ & \textit{74.84}$^\star$ & \textit{53.10}$^\star$& \textit{90.98}$^\star$ & \textit{97.10} & \textit{\textbf{39.66}}$^\star$& \textbf{\textit{78.86}}$^\star$ & \textit{91.13}$^\star$ & \textit{\textbf{46.59}}$^\star$& \textit{82.64}$^\star$ & \textit{96.12}$^\star$ & \textit{29.59}$^\star$& \textit{67.21}$^\star$ & \textit{85.34}$^\star$ & \textit{4.51 / 45.06}\\
\bottomrule
\end{tabular}
\end{adjustbox}
\end{table*}

Unlike existing datasets captured from broadcast angles with professional equipment, TTA videos are recorded by sports analysts using handheld cameras during elite competitions (Paralympics, World Para Future). This realistic setup introduces frequent occlusions, motion blur, and dynamic viewpoints—conditions representative of practical deployment scenarios. Moreover, as shown in Figure~\ref{fig:event-density}, TTA exhibits significantly higher event density (7.73 events per 100 frames) compared to Tennis (3.81), FineDiving (2.34), and Figure Skating (1.47). This dense event distribution within narrow temporal windows demands fine-grained temporal modeling capabilities, making TTA a particularly challenging benchmark for evaluating precise event spotting methods.

\subsection{Evaluation and Implementation Details}
MFS is evaluated as a drop-in replacement for GSM using the E2E-Spot framework \cite{hong2022spotting}.  
Models are trained on 100-frame clips with AdamW \cite{loshchilov2017decoupled} (lr=$1\times10^{-3}$, cosine decay, 3 warm-up epochs) for 50 epochs, batch size 8.  
RegNetY-200 \cite{radosavovic2020designing} and ResNet-18 \cite{he2016deep} serve as backbones.  Performance is reported in mean Average Precision (mAP) across all classes under $\delta\!\in\!\{0,1,2\}$, where $\delta\!=\!0$ requires exact-frame matching and $\delta\!=\!1,2$ allow $\pm1/\pm2$-frame tolerance.  
All experiments are conducted on a single NVIDIA L40S GPU.

\subsection{Comparison with State-of-the-Art}
\label{sec:sota_comparison}

We evaluate MFS against six PES models spanning two paradigms. The transformer-based category includes ASTRA \cite{xarles2023astra} and ViT \cite{dosovitskiy2020image}. The CNN-based category includes E2E-Spot \cite{hong2022spotting}, T-DEED \cite{xarles2024t}, UGLF \cite{tran2024unifying}, and ASTRM \cite{santra2025precise}. E2E-Spot and UGLF utilize GSM \cite{sudhakaran2020gate}, while T-DEED employs GSF \cite{sudhakaran2023gate}. For CNN-based methods, we evaluate both the original temporal module and MFS as a drop-in replacement.

\textbf{Standardized Evaluation Protocol.}
To rigorously evaluate the effectiveness of the \textit{temporal mixing architectures} in isolation, we standardized all methods to a RegNetY-200 \cite{radosavovic2020designing} backbone, a fixed 100-frame context window, and a standard cross-entropy loss schedule. This decouples the temporal module's performance from auxiliary engineering factors (e.g., specialized loss functions or heavy backbones).

\textbf{Main Results and Efficiency.}
Table~\ref{tab:main-results} presents results across five datasets and three temporal tolerances. Under the standardized evaluation protocol, MFS demonstrates broad superiority over existing temporal modules.
Quantitatively, MFS achieves the best reported performance in \textbf{12 out of 15} dataset-tolerance combinations (columns).
When comparing individual model-backbone configurations (e.g., MFS vs. GSM, MFS vs. GSF across all settings), MFS achieves the winning score in \textbf{36/45} direct comparisons.
This dominance is particularly evident against lightweight baselines, where MFS improves over GSM by +4.09\% mAP on Tennis ($\delta=0$) and +3.08\% on TTA ($\delta=1$).
Comparison against the heavy SOTA baseline ASTRM \cite{santra2025precise} further highlights the strength of MFS. While the standardized ASTRM achieves 55.31\% mAP on Tennis ($\delta=0$), MFS matches this performance (55.51\%) without requiring the heavy computational overhead.
Specifically, ASTRM requires 67.24 GFLOPs, whereas MFS operates at only 45.06 GFLOPs. Thus, MFS delivers \textbf{SOTA-level accuracy with 33\% fewer FLOPs} (1.5$\times$ efficiency gain), proving that complex temporal mechanisms are not required for top-tier performance when efficient multi-scale shifting is applied.

\textbf{Robustness on Challenging Data.}
While MFS matches ASTRM on standard benchmarks, it demonstrates superior robustness on the proposed TTA dataset, which features higher event density and frequent occlusions. On TTA ($\delta=1$), MFS achieves \textbf{69.50\%}, significantly outperforming the standardized ASTRM (64.86\%) by \textbf{+4.64\%}. This indicates that MFS's combination of multi-scale shifting and grouped focusing handles rapid, complex actions more effectively than the heavier baseline when computational resources are equalized.

\textbf{Generalization Across Architectures.}
To assess versatility, we test MFS with ResNet-18 \cite{he2016deep} (results in Supp.~§3.1). MFS provides consistent gains for E2E-Spot (14/15 comparisons) and UGLF (13/15 comparisons), confirming it is a highly effective upgrade for standard CNN backbones.

\begin{table}[t]
\centering
\caption{Ablation study on attention head count, temporal distance range, and clip length using the MFS module on the TTA dataset. All results are reported in mAP (\%), default clip length=100, best result is in bold. }
\label{tab:ablation}
\renewcommand{\arraystretch}{1.0}
\scriptsize
\resizebox{0.8\linewidth}{!}{%
\begin{tabular}{l l | c c | c}
\toprule
\textbf{Setting} & \textbf{Value} & $\delta=1$ & $\delta=2$ & \textbf{P(M) / GFLOPs} \\
\midrule
\multicolumn{5}{l}{\textit{Multi-Scale Shift}} \\
E2E-GSM & - & 66.42 & 76.90 & 4.45 / 42.18 \\
Distance & {[1,2]} & \textbf{66.08}& \textbf{77.31}& 4.47 / 43.26 \\
Distance & {[1,2,3]} & 65.82 & 74.67 & 4.49 / 44.32 \\
Distance & {[1--4]} & 64.39 & 74.91 & 4.51 / 45.40 \\
\midrule
\multicolumn{5}{l}{\textit{Group Focused Module}} \\
E2E-GSM & - & 66.42 & 76.90 & 4.45 / 42.18 \\
Groups & 1 & \textbf{67.02} & \textbf{75.74} & 4.46 / 42.56\\
Groups & 2 & 65.39 & 72.35 & 4.47 / 42.92 \\
Groups & 3 & 64.01 & 71.94 & 4.47 / 43.28 \\
\midrule
\multicolumn{5}{l}{\textit{Varying Number of Groups (Distance = [1,2,3])}} \\
Groups & 1 & 65.99 & 75.38 & 4.50 / 44.68 \\
Groups & 2 & \textbf{69.50}& \textbf{78.12}& 4.51 / 45.06 \\
Groups & 3 & 65.87 & 75.87 & 4.51 / 45.42 \\
Groups & 4 & 61.01 & 72.98 & 4.52 / 45.78 \\
\midrule
\multicolumn{5}{l}{\textit{Varying Temporal Shift Distance (Groups = 2)}} \\
Distance & {[1,2]} & 64.78 & 74.10 & 4.48 / 43.98 \\
Distance & {[1,2,3]} & \textbf{69.50}& \textbf{78.12}& 4.51 / 45.06 \\
Distance & {[1--4]} & 67.54 & 71.59 & 4.53 / 46.12 \\
Distance & {[1--5]} & 64.18 & 67.41 & 4.55 / 47.18 \\
Distance & {[1--6]} & 64.03 & 75.38 & 4.57 / 48.26 \\
\midrule
\multicolumn{5}{l}{\textit{Varying Clip Length with Distance=[1,2,3], Groups=2}} \\
Length & 50 & 58.56  & 66.80 & 4.51 / 21.46 \\
Length & 100 & \textbf{69.50} & \textbf{78.12} & 4.51 / 45.06 \\
Length & 150 & 62.24 & 74.50 & 4.51 / 64.22 \\
Length & 200 & 57.29 & 66.78 & 4.51 / 85.50 \\
\bottomrule
\end{tabular}}
\end{table}

\subsection{Ablation Study}
\label{sec:ablation_study}
We perform ablations on the TTA dataset to validate the contributions of multi-scale temporal shifting and grouped spatial focusing.
\textbf{Component Analysis.}  
Table~\ref{tab:ablation} (top) isolates each component. Multi-scale shifting alone provides unstable gains, as larger temporal windows introduce background noise. Grouped focus alone improves spatial selectivity but offers limited benefit without extended temporal context. Combining both yields the strongest improvement (+3.08 mAP), confirming that spatial grouping identifies salient regions while multi-scale shifts propagate them effectively over time.
\textbf{Hyperparameter Effects.}  
Table~\ref{tab:ablation} (middle) varies group count $G$ and temporal distances. Increasing from $G=1$ to $G=2$ gives the largest gain, enabling specialization on different spatial regions, while $G>2$ over-fragments channels and hurts stability. For temporal ranges, $\{1,2,3\}$ offers the best balance between short-term precision and extended context; shorter ranges reduce receptive field, while overly large ones introduce irrelevant frames. We adopt $G=2$ and distances $\{1,2,3\}$ as defaults.
\textbf{Clip Length.}  
Following \cite{hong2022spotting}, we test clip lengths $L\in\{50,100,150,200\}$. As shown in Table~\ref{tab:ablation} (bottom), $L=100$ performs best—short clips lack context, while longer clips add cost without accuracy gains. All main results use $L=100$.

%% file: sec/6_conclusion.tex
\section{Conclusion}



We introduced MFS, a lightweight enhancement of GSM that expands temporal context and adds spatial selectivity through multi-scale shifts and grouped focusing. With only 1.3\% additional parameters, MFS consistently improves mAP across five PES benchmarks while remaining highly efficient.  
We also presented the TTA dataset, the first table-tennis PES benchmark with realistic handheld recordings and high event density.  
Overall, our results show that targeted improvements to lightweight modules can match or surpass more complex alternatives while maintaining practicality for real-world deployment.